\documentclass[10pt,twocolumn,letterpaper]{article}

\usepackage{cvpr}              

\usepackage{graphicx}
\usepackage{amsmath}
\usepackage{amssymb}
\usepackage{booktabs}
\usepackage{xspace}

\usepackage[breaklinks,colorlinks]{hyperref}

\usepackage{xcolor}
\newcommand{\GITHUBLINK}{\href{http://noise-aware-nerf.github.io}{noise-aware-nerf.github.io}}

\usepackage{soul}

\def\cvprPaperID{8196} 
\def\confName{CVPR}
\def\confYear{2022}

\begin{document}

\title{NAN: Noise-Aware NeRFs for Burst-Denoising}


\author{Naama Pearl, Tali Treibitz\\
Dept. of Marine Technologies, School of Marine Sciences\\
University of Haifa, Israel\\
{\tt\small \{npearl@campus,ttreibitz@univ\}.haifa.ac.il}
\and
Simon Korman\\
Dept. of Computer Science\\
University of Haifa, Israel\\
{\tt\small skorman@cs.haifa.ac.il}
}
\twocolumn[{%
\renewcommand\twocolumn[1][]{#1}%
\maketitle

\begin{center}
    \centering
    \captionsetup{type=figure}
    \vspace{-0.26in}
    \includegraphics[width=0.99\textwidth]{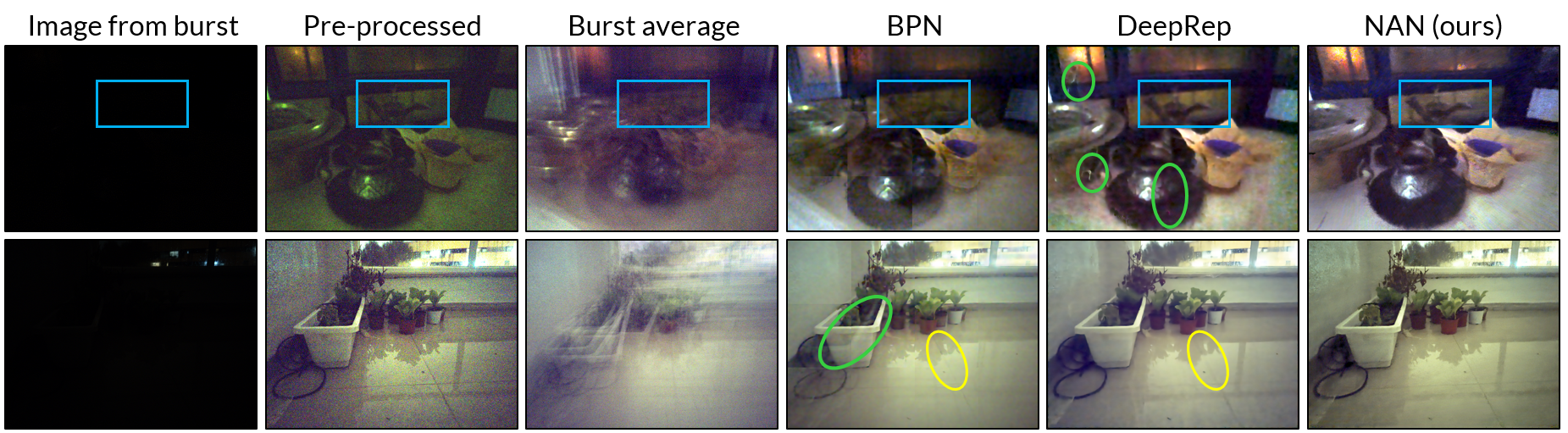} \vspace{-0.08in}
    \captionof{figure}{\textbf{Burst denoising in challenging real-world low-light scenes.}  
    The dark burst images, scaled before processing, contain high levels of noise and significant camera motion that can be seen in their averages.  
    The results of both BPN~\cite{xia2020basis} and DeepRep~\cite{bhat2021deep} are generally more blurry and lack detail compared to those of NAN, the proposed method. Blue rectangles mark a clear example for comparison. Green and yellow ellipses show artifacts and missing detail correspondingly in competitor results. \textbf{The reader is encouraged to zoom-in.}} 
    \label{fig:real_world}\vspace{-1pt}
\end{center}%
}]

\begin{abstract}
\vspace{-0.08in}
Burst denoising is now more relevant than ever, as computational photography helps overcome sensitivity issues inherent in mobile phones and small cameras.
A major challenge in burst-denoising is in coping with pixel misalignment, which was so far handled with rather simplistic assumptions of simple motion, or the ability to align in pre-processing. Such assumptions are not realistic in the presence of large motion and high levels of noise. 

We show that Neural Radiance Fields (NeRFs), originally suggested for physics-based novel-view rendering, can serve as a powerful framework for burst denoising. 
NeRFs have an inherent capability of handling noise as they integrate information from multiple images, but they are limited in doing so, mainly since they build on pixel-wise operations which are suitable to ideal imaging conditions. 

Our approach, termed NAN\footnote{Refer to the project website: \texttt{\GITHUBLINK}}, leverages inter-view and spatial information in NeRFs to better deal with noise. It achieves state-of-the-art results in burst denoising and is especially successful in coping with large movement and occlusions, under very high levels of noise. With the rapid advances in accelerating NeRFs, it could provide a powerful platform for denoising in challenging environments. 

\end{abstract}


\vspace{-0.3cm}
\section{Introduction}
\label{sec:intro}


Burst denoising has become the de-facto method for low-light imaging, especially in handheld mobile devices that perform onboard processing~\cite{hasinoff2016burst,liba2019handheld}. 
It is built on capturing multiple short-exposure (dark and noisy) frames, which are then integrated into a single coherent image. 
The main challenge of compensating for motion and occlusion, especially under high noise, is evident in limitations of current applications (e.g., mobile phone apps ask to ``hold still'' while capturing in night mode).
Dealing with large parallax could significantly enhance denoising, by allowing the use of longer bursts than (the typical 8) currently common in cameras, and might enable more flexible imaging during movement (e.g., from a vehicle).  

Recently,  methods based on Neural Radiance Fields (NeRFs) have been demonstrated to be powerful in rendering novel views of scenes, enabling the synthesis of intricate details due to reflections and occlusions, yet all  these methods consider clean high-resolution images as inputs.
Since NeRFs integrate information from multiple images, they have strong potential to be leveraged in multi-frame image restoration tasks - applications they were not designed for.
We show in this paper that they can be very powerful for burst denoising. 


NeRFs serve as implicit scene priors and thus can inherently handle noise~\cite{du2021neural,mildenhall2022nerf}. 
However, when using a small number of input images this capability is limited, since current architectures operate separately on each pixel with local computations that are highly prone to noise - implying much room for improvement.
We demonstrate that by adding inter-view and spatial awareness to the network we significantly improve its ``noise-awareness'' and produce SOTA results in burst denoising, especially under large motion and high noise (see Fig.~\ref{fig:real_world}). 
To avoid specific per-scene training, we build on the recently proposed IBRNet~\cite{wang2021ibrnet} that pre-trains on different image sets and is able to produce novel views during inference time in new unseen scenes, using as few as $8$ images.

To summarize our contributions:
(i) We achieve SOTA results in burst denoising; 
(ii) We successfully exploit the natural power of radiance fields as scene priors by augmenting them with novel noise-awareness components, both in the spatial and cross-view domains;
(iii) We demonstrate the advantages of our approach (NAN) over SOTA burst-denoising methods, which operate in the image plane, while NAN, being NeRF based, explicitly works in 3D space - imperative for dealing with large motion and high noise. 







\section{Related work}
\label{sec:related}

\subsection{Neural Radiance Fields}

The field of Neural Rendering has seen a surge of interest since the recent appearance of the NeRF~\cite{mildenhall2020nerf} representation for image synthesis that has promoted rapid improvements in many aspects of the problem (see~\cite{tewari2020state} for a comprehensive review on the SOTA prior to the introduction of NeRFs and~\cite{tewari2021advances} for the current SOTA).

The seminal paper~\cite{mildenhall2020nerf}  presents a    volume rendering approach, in which a neural network in the form of a multi-layer perceptron (MLP) encodes an implicit volumetric representation of a 3D scene. It is trained in a supervised manner from a set of posed images, to serve as a 5D radiance field that provides volume density and view-dependent radiance as a function of 3D location and 2D viewing direction.

One active line of followups~\cite{garbin2021fastnerf,mueller2022instant,reiser2021kilonerf,yu2021plenoctrees,hedman2021baking} focuses on improving performance, with accelerations of several orders of magnitude in rendering speed, allowing for application in real-time scenarios. 
Another  line of work, including Bundle-Adjusting~\cite{lin2021barf} and Self-Calibrating~\cite{jeong2021self} NeRFs, performs joint learning of camera pose registration and 3D neural representation, allowing NeRFs to be applicable to a much wider range of setups.

One of the other main challenges has been in relaxing the assumption made by~\cite{mildenhall2020nerf} regarding the input scene being photometrically static $-$ that the density and radiance of the world are constant. A large collection of works deals with non-static \textit{density}, i.e., the presence of motion in the scene. DNeRF~\cite{pumarola2021d}, Nerfies~\cite{park2021nerfies} and NeRFlow~\cite{du2021neural} do so by giving a full 4D representation  
that fully accommodates the motion in the scene and can create novel video clips from a customizable camera capturing motion paths.

The `NeRF in the Wild' (NeRF-W) work~\cite{martin2021nerf} treats motion differently, by explaining it out, in order to reconstruct the static scene that is shared by the input images. 
It also deals with non-static \textit{radiance} that is due to different per-camera exposure, color correction and tone-mapping, to avoid inaccurate reconstructions. 
This is done by modeling image-dependent radiance using a per-image latent embedding vector, which is suitable for modeling \textit{global} photometric phenomena, but not independent pixel-level inconsistencies that stem from noise. The approach we present deals directly with such local inconsistencies, by exploiting cross-image and spatial within-image information. 

We build on the architecture of {IBRNet}~\cite{wang2021ibrnet} - a differentiable image-based rendering network, which similarly to {pixelNeRF}~\cite{yu2021pixelnerf} can generalize to new scenes, represented by possibly only a few images, without the need for test-time optimization like other prior works that are trained to model a \textit{specific} scene. 
To the best of our knowledge, NAN is the first NeRF-based work that explicitly deals with significant photometric noise. 
Concurrently with our work, `Nerf in the Dark'~\cite{mildenhall2022nerf} demonstrate the power of NeRFs in generating novel clean views using many dozens of extremely low light linear images. In the NAN framework, we make use of the noisy target image and demonstrate efficient denoising using as little as 8 frames.



\subsection{Burst Denoising}


Recent works~\cite{mildenhall2018burst,xia2020basis,kokkinos2019iterative,rong2020burst} have demonstrated that burst denoising can overcome many inherent problems of long exposure photography, including motion blur and non-uniform dynamic ranges across an image that result in dark-and-noisy or overexposed regions.


Deep Burst Denoising~\cite{godard2018deep} uses a recurrent neural network to denoise a burst after stabilizing it using a Lucas-Kanade tracker to find correspondences between successive frames, followed by a rotation-only motion model to estimate a homography between the frames. In Kernel Prediction Networks~\cite{mildenhall2018burst} a CNN is trained to predict per-pixel per-input-image specific 2D denoising kernels, which are used as 3D blending weights to obtain a clean target image from a noisy burst. However, they allow motion of up to $2$ pixels and in practice - a vast majority of the blending weights are assigned to the target-image kernels, which means that the use of signal information from the other images is sub-optimal. Basis Prediction Networks~\cite{xia2020basis} enable handling larger motions by using kernels of a much larger spatial extent, which is made possible by representing them as linear combinations of a small set of basis elements. 

Very recently, SOTA burst-denoising results were presented by DeepRep~\cite{bhat2021deep}, which proposes transforming the MAP estimator used in image restoration tasks into a deep feature space. They tackle the motion issue by aligning each input image to the target image using optical flow~\cite{sun2018pwc}. This alignment approach is possibly sufficient for small motion and subtle noise, but is prone to errors in the presence of higher noise, and unlike the NeRF based approach that we adopt - it only considers a pair of frames at a time without reasoning about the scene's underlying 3D structure.

Taking into account the 3D structure enables coping with very high levels of noise. 
Recent burst denoising works~\cite{bhat2021deep,mildenhall2018burst,xia2020basis} reported results on noise levels of  a standard deviation of $\sim0.2$ for an image in the range $[0,1]$, while others~\cite{rong2020burst,du2021neural} test on levels of $\sim 0.1$. 
However, higher noise levels like the ones we consider in this work of up to $\sim0.4$ are frequent in low-light~\cite{wang2021seeing} settings, especially in non-uniformly lit scenes, haze conditions, and underwater.



%



%
%
%



\section{Background}
\label{sec:back}

\subsection{Problem Setup}\label{sec:burst-denoising setup}

In the problem of \textit{burst denoising}, given a burst of $N$ noisy images of a scene $\{I_n\}_{n=1}^N$, the goal is to generate a clean version of one of the images by jointly processing the entire burst. 
In this work, we focus on bursts with large camera motion between frames and do not assume consecutive frames to be closely aligned. Hence we treat the burst as an un-ordered set rather than a sequence.

\begin{figure} [t]
    \centering \vspace{-2pt}
    \includegraphics[width=0.75\columnwidth]{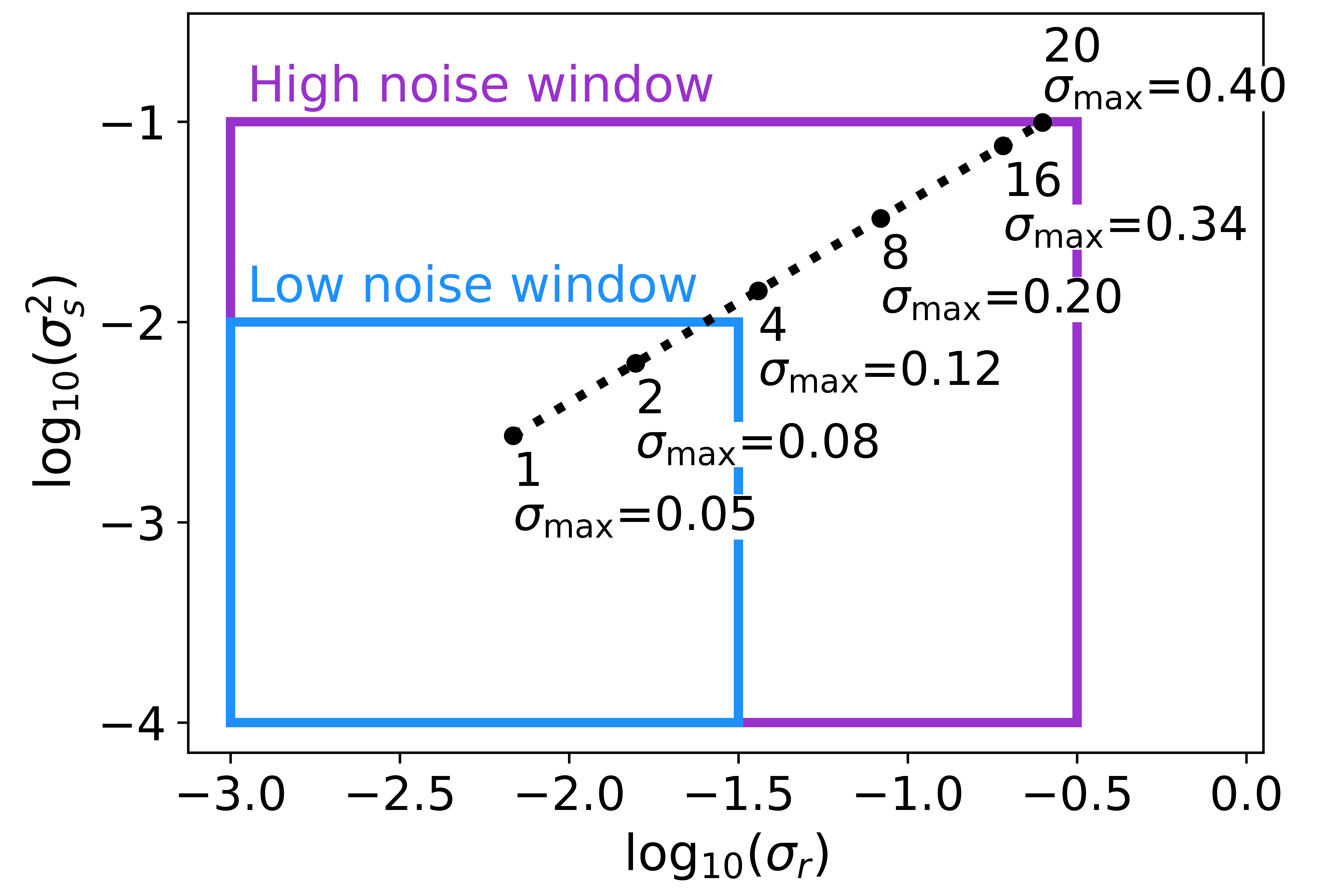}\vspace{-0.2cm}
    \caption{\textbf{Noise levels} used for training and testing. The value of $\sigma_{\max}$ indicates the maximum noise level in the image, relative to a maximum image intensity of $1$. See text for details. 
    } \vspace{-5pt}
    \label{fig:noise_levels}
\end{figure}

We follow the noise model used in
\cite{mildenhall2018burst,xia2020basis,bhat2021deep} where the noisy version of a clean linear image  $I_n^c$ is given by:
\begin{equation}\label{eq:noise}
{I_n}(x) \sim \mathcal{N}\Big(I^c_n(x), \sigma_r^2 + \sigma_s^2I^c_n(x)\Big)\;\;,
\end{equation}
\noindent where $x$ is an image coordinate, $\sigma_r$ is the signal independent read-noise parameter, $\sigma_s$ is the signal dependent shot-noise parameter and $\mathcal{N}$ represents the Gaussian distribution. 
For convenience, following~\cite{mildenhall2018burst,xia2020basis,bhat2021deep}, we report results on \textit{gain} levels of an example camera (see Appendix A in~\cite{mildenhall2017burst_ARXIV} for a detailed explanation on the connection between sensor gain and noise parameters).
These works performed model training in a relatively low-noise region (the blue rectangle in Fig.~\ref{fig:noise_levels}, upper limited by a gain of around $4$), with evaluation on gain levels $1$, $2$, $4$ and $8$.
As we want our method to operate in higher levels of noise, we extend the training range up to gain $20$ (purple rectangle in Fig.~\ref{fig:noise_levels} with a maximum noise level equivalent to a standard deviation of $\sim 0.4$ for images in [0,1]) and add additional evaluation gain levels in the extended range (black points in~\ref{fig:noise_levels}).
For fair comparison, we retrained \cite{bhat2021deep} and \cite{xia2020basis} on the same noise region.

\begin{figure*}[t]
    \centering\vspace{-8pt}
    \includegraphics[width=1\textwidth]{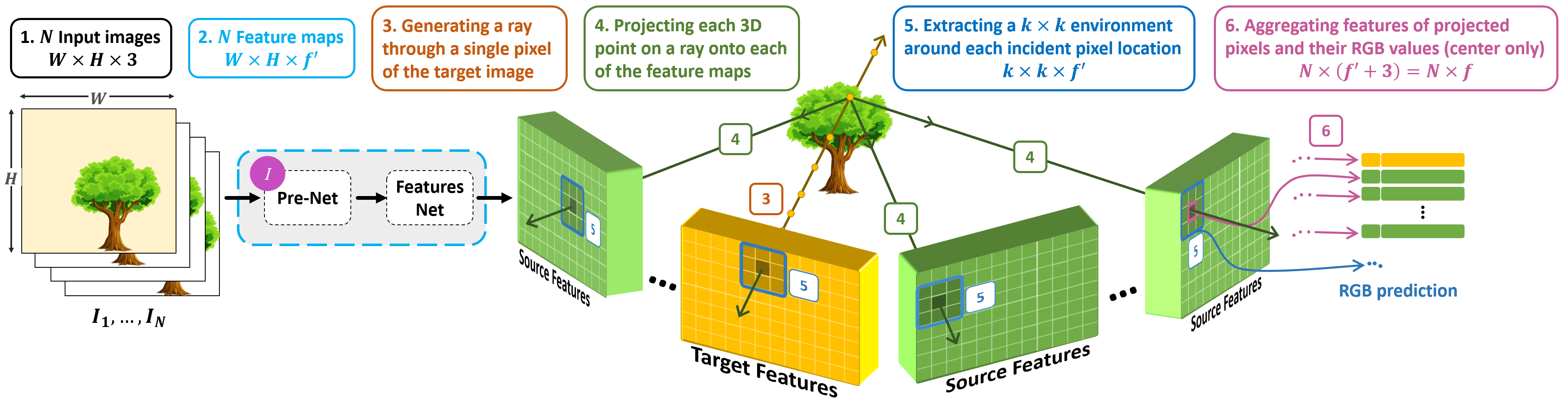}\vspace{-0.2cm}
    \caption{\textbf{Our IBRNet-based architecture - part 1: Feature extraction and ray projection}. We train a Pre-Net layer (\emph{I}) to process the pixels before entering a feature extraction network. Then, for each pixel in the target image (yellow) a ray is projected. Each point sample along the ray is projected onto each of the views. The features from all viewing directions are then aggregated to enter the main network.}
    \label{fig:ray_scheme}
\end{figure*}


\begin{figure*}[t]
    \centering\vspace{-8pt}
    \includegraphics[width=1\textwidth]{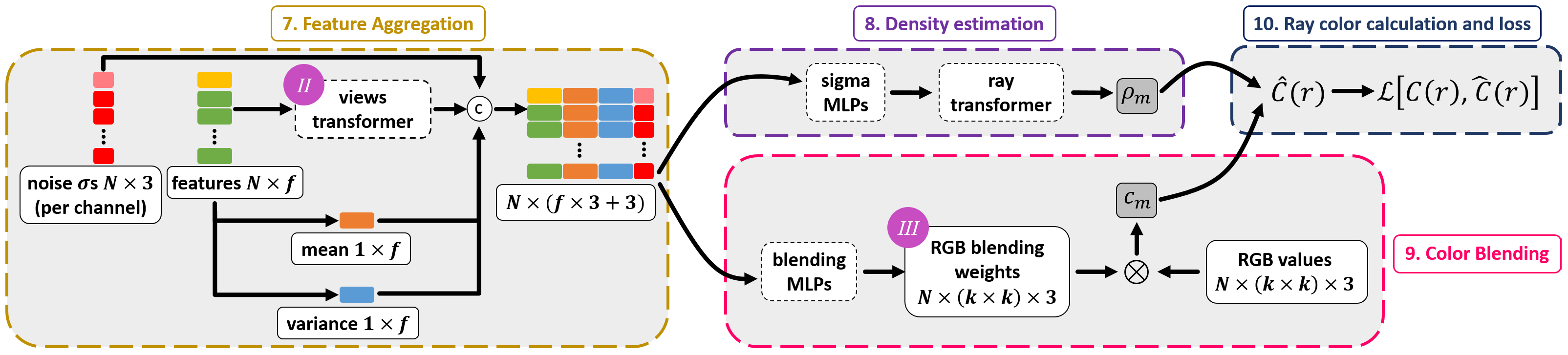}
    \vspace{-16pt}
    \caption{\textbf{Our IBRNet-based architecture - part 2: Density and color estimation}. 
   The ray projection features are the input to the {\color{brown} feature aggregation} stage of IBRNet~\cite{wang2021ibrnet}, which we extend with per-pixel noise-level inputs and a views transformer (\emph{II}) to enrich the relative representation. The aggregated features are inputted into two parallel networks: {\color{violet} Density Estimation} that compares the features and determines if they stem from the same object point, in the form of an output density $\rho_m$; {\color{magenta} Color Blending} that calculates the pixel blending weights, where we expand the weights to include a window around the pixel and separate the weights for the RGB channels (\emph{III}).}
    \label{fig:density_color_scheme}\vspace{-5pt}
\end{figure*}

\subsection{IBRNet}

The NAN network builds upon the scheme and architecture of IBRNet~\cite{wang2021ibrnet}. Among alternative NeRF-based methods, we chose it as our baseline, first and foremost, since it pre-trains on an entirely separate set of scenes from the test-time scene on which it does not need to optimize, therefore having a relatively short inference time, requiring only a small number of images. In addition, due to its properties, common to all NeRFs, of being permutation invariant (with respect to input image ordering) and being able to handle a variable number of input images (just like~\cite{kokkinos2019iterative,rong2020burst,bhat2021deep}, but unlike burst denoising methods like~\cite{mildenhall2018burst,xia2020basis}).

%


The general structure of IBRNet~\cite{wang2021ibrnet} is summarized in Figs.~\ref{fig:ray_scheme} and~\ref{fig:density_color_scheme}. Each pixel in the novel-view `target' image is processed independently. For each such pixel, a ray $r$ is projected into 3D space in order to estimate the existence and appearance of objects in the real-world along this ray. To achieve that, the ray is sampled (regularly on an inverse-depth scale) at a sequence of 3D positions indexed by $m=1,\ldots,M$ (Fig.~\ref{fig:ray_scheme}). 
For each such 3D point, the set of pixels imaging it in the other views, along with their local image features, are accumulated, while concatenating the mean and variance vectors as they are important signals for the downstream density and color prediction (Fig.~\ref{fig:density_color_scheme} step $7$). 
These representations
are processed to predict the point color $c_m$, and its density $\rho_m$ (Fig.~\ref{fig:density_color_scheme}).

The density is used to calculate the probability $w_m$ that an object exists in that particular 3D location:
\begin{equation} \label{eq:alpha}
   w_m=\left(1-e^{-\rho_m}\right) /\; e^{\left(\sum_{j=1}^{m-1}\rho_j\right)}\;.\;\;
\end{equation}
Eventually, these weight-color pairs, computed for each sample location along the ray, are  integrated to produce the final color of the predicted target image pixel $\hat{C}(r)$:
\begin{equation}
 \hat{C}(r)=\sum_{m=1}^{M}w_m c_m \;\;.
\end{equation}

This process is repeated in two stages, `coarse' and `fine', where the sampling is refined in the second stage around the depth values that most influence the rendering (i.e., at visible surfaces). 
In training, the novel pose is taken to be one of the available (non-input) viewpoints, allowing for a loss that measures the rendering quality by comparing the original and rendered pixel colors.


\section{Method}

NeRFs, and in particular IBRNet, as methods for novel-view synthesis were designed with a focus on dealing with the intricate combination of significant camera motion and complex 3D scene geometry. The burst-denoising setup is different in two main aspects: (i) It requires working with degraded noisy images (\eg, captured in low-light conditions), which makes things more challenging; (ii) It does not require generating novel views, but rather fixing an existing view image, which is a valuable source of data that NeRFs do not possess.

\subsection{A Simple Baseline} \label{sec:simple_baseline}

As a  baseline for our method we simply train the original IBRNet described above on a dataset of noisy images, and denote the resulting network IBRNet-\emph{N}. 
For doing so we created a new dataset which we term \mbox{LLFF-$N$}, by taking (clean) scene images from the LLFF~\cite{mildenhall2019local} and NeRF~\cite{mildenhall2020nerf} datasets, which were used by IBRNet~\cite{wang2021ibrnet}, and adding varying levels of noise according to the model in Eq.~\eqref{eq:noise}. 
Since this model refers to \textit{linear} images, we first ``linearize" the images by applying inverse gamma correction and inverse random white balancing as in~\cite{rong2020burst,xia2020basis,bhat2021deep} before we add the noise. These pseudo-linear images are the input to the tested methods, where one of the scene images is chosen at random as the target to be denoised (against which loss is calculated) and the output linear denoised image is reprocessed by applying the respective gamma correction and white balancing. 
Although the noise in LLFF-\emph{N} is simulated, its inter-frame motion is real and hence realistic in terms of occlusions and projections, in contrast with the random image-plane translations used in~\cite{mildenhall2018burst,xia2020basis}.

Examining the denoising performance of IBRNet and of IBRNet-\textit{N}, its version retrained-with-noise on the LLFF-$N$ dataset, we observe promising initial results. Nevertheless, it is clear there is room for improvement. 
In the following, we detail our main proposed adaptations over the baseline model, which extend IBRNet to make it significantly more capable of handling image noise, outperforming SOTA burst denoising methods (see Sec.~\ref{sec:results}).

In particular, the quality of a radiance field depends critically on its ability to estimate both the density $\rho_m$ and the color $c_m$ associated with each sampled 3D position along the projected rays. Note that the density and color are learned jointly, in an end-to-end manner in which the ability to estimate one clearly affects the ability to estimate the other. We indeed observe a degradation in both as the baseline is not truly suited to handle noise.

In Sec.~\ref{sec:density_estimation}, we focus on improving the density prediction of the network by a specific change in the feature extraction sub-network (purple \emph{I}  in Fig.~\ref{fig:ray_scheme}, Sec.~\ref{sec:feat_extract}) and by modifying the main feature aggregation stage (purple \emph{II} in Fig.~\ref{fig:density_color_scheme}, Sec.~\ref{sec:feat_aggr}).
In Sec.~\ref{sec:color_blend}, we describe the enhancement of the color predictions of the network by using specific spatial considerations that are suitable for color blending in noisy regimes (purple \emph{III} in Fig.~\ref{fig:density_color_scheme}).
%

\subsection{Noise-Aware Density Estimation } \label{sec:density_estimation}

\begin{figure}[t!]
    \centering
    \includegraphics[width=0.99\columnwidth]{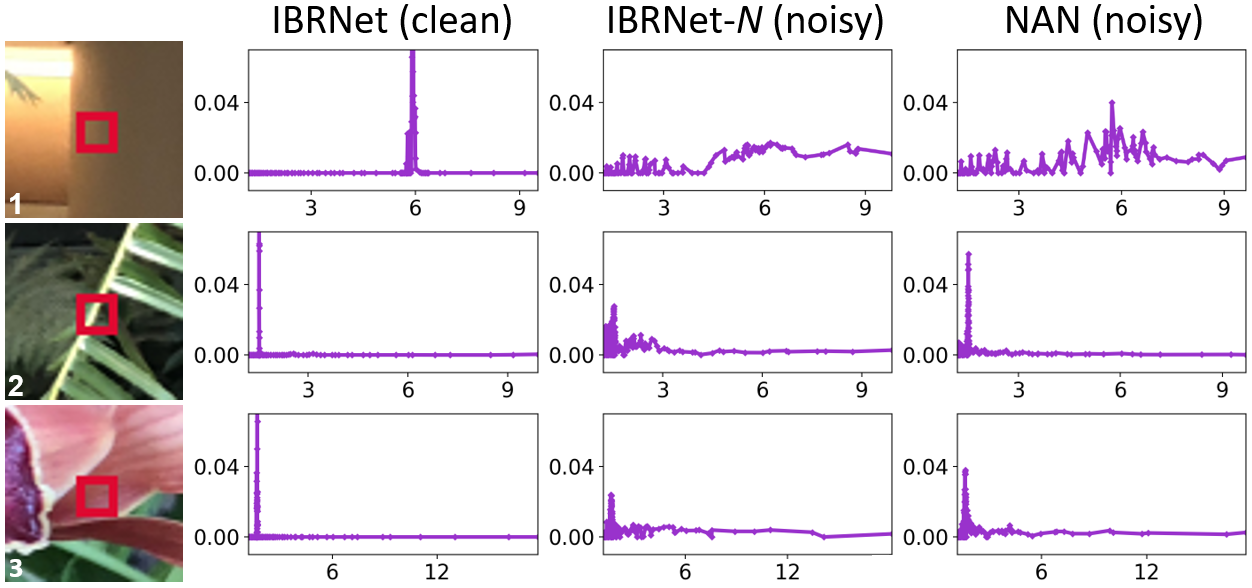}\vspace{-0.2cm}
    \caption{Ray density weight ($w_m$) plot for several object points. In clean images (\textbf{left}) the distributions are narrow, resembling a $\delta$ function that represents the true surface depth. 
    In IBRNet-\emph{N} (\textbf{center}), the $\delta$ shape is replaced by widely spread densities. NAN (\textbf{right}) better handles the noise, evident in more density accumulating around the original $\delta$ location, resulting in better details in the restored image. In smooth regions, such as in row 1, there is an advantage in a wider distribution that enables improved denoising.} \vspace{-5pt}
    \label{fig:sigma_plots}
\end{figure}

Fig.~\ref{fig:sigma_plots} depicts the ray weights $w_m$ (directly related to the density through Eq.~\eqref{eq:alpha})  for several target image points. For sharp rendering, we expect the weight function to be close to a $\delta$ function around the true depth, as is the case in clean images (Fig.~\ref{fig:sigma_plots}~left). This ideal situation degrades with the addition of noise. 
In the IBRNet-$N$ case (Fig.~\ref{fig:sigma_plots}~middle) the distribution is very wide and less deterministic regarding the object's location. The spread of the distribution can be explained by the difficulty to accurately predict the true depth, but perhaps also as something that the learning resorts to in the optimization, since spread-out density results in photometric blurring which reduces noise (at the cost of losing details). 
The following suggested adaptations are shown to improve the density distribution predictions.

\vspace{-4pt}
\subsubsection{Feature Extraction} \label{sec:feat_extract}


We observed that the feature extraction network used in~\cite{wang2021ibrnet} does not generalize well to handle noisy inputs, since we were able to improve performance by na\"ively pre-processing the noisy images individually using even a simple Gaussian low-pass filter. 
We found the original entry-point convolutional layer incapable of performing simple noise filtering, due to its large $7 \times 7$ kernel size and immediate spatial resolution reduction by a factor of two.

Motivated by these observations, we suggest a  simple addition to the truncated ResNet34 encoder-decoder network used in IBRNet~\cite{wang2021ibrnet}, in the form of an additional (trainable)
single convolution layer  with $3 \times 3$ kernel size, $3$ output channels and without an activation function, at the entry point of the network (see purple~\emph{I} in Fig.~\ref{fig:ray_scheme}). Our new `Pre-Net' layer, whose weights we initialize to Gaussian per-channel filters, preserves the spatial and channel dimension of the input image and reduces the input noise efficiently. It impacts the entire denoising performance, as demonstrated in our detailed ablation study (Sec.~\ref{sec:ablation}).
 
Note that the RGB values of the source images are used twice in the architecture (in addition to being the input of the feature embedding): first, they are concatenated to the corresponding feature vectors (see Fig.~\ref{fig:ray_scheme} step 6) and second, as the input to the final blending (see Fig.~\ref{fig:density_color_scheme} step 9). In experimentation, we found that better performance is achieved when passing the `Pre-Net'-filtered values in the first case and the original noisy pixels values in the second. Hence, overall, this solution is more accurate and adaptive compared to the alternative of simple input noise filtering.




\vspace{-4pt}
\subsubsection{Feature Aggregation} \label{sec:feat_aggr}

Recall that the pixel feature embedding vectors are the main input of the density and color estimation parts of the network  (Fig.~\ref{fig:density_color_scheme}). 
On their entry, they are each expanded by concatenation to include their cross-view mean and variance feature vectors of the specific pixel (step 7).
We examine this extended feature, by separately considering the original and extended statistics (mean and variance) parts. 
Their relative contribution to the final predictions can be  ablated by  zeroing their part in  the vector at inference.

\begin{figure}[t]
    \centering\vspace{-8pt}
    \includegraphics[width=0.99\columnwidth]{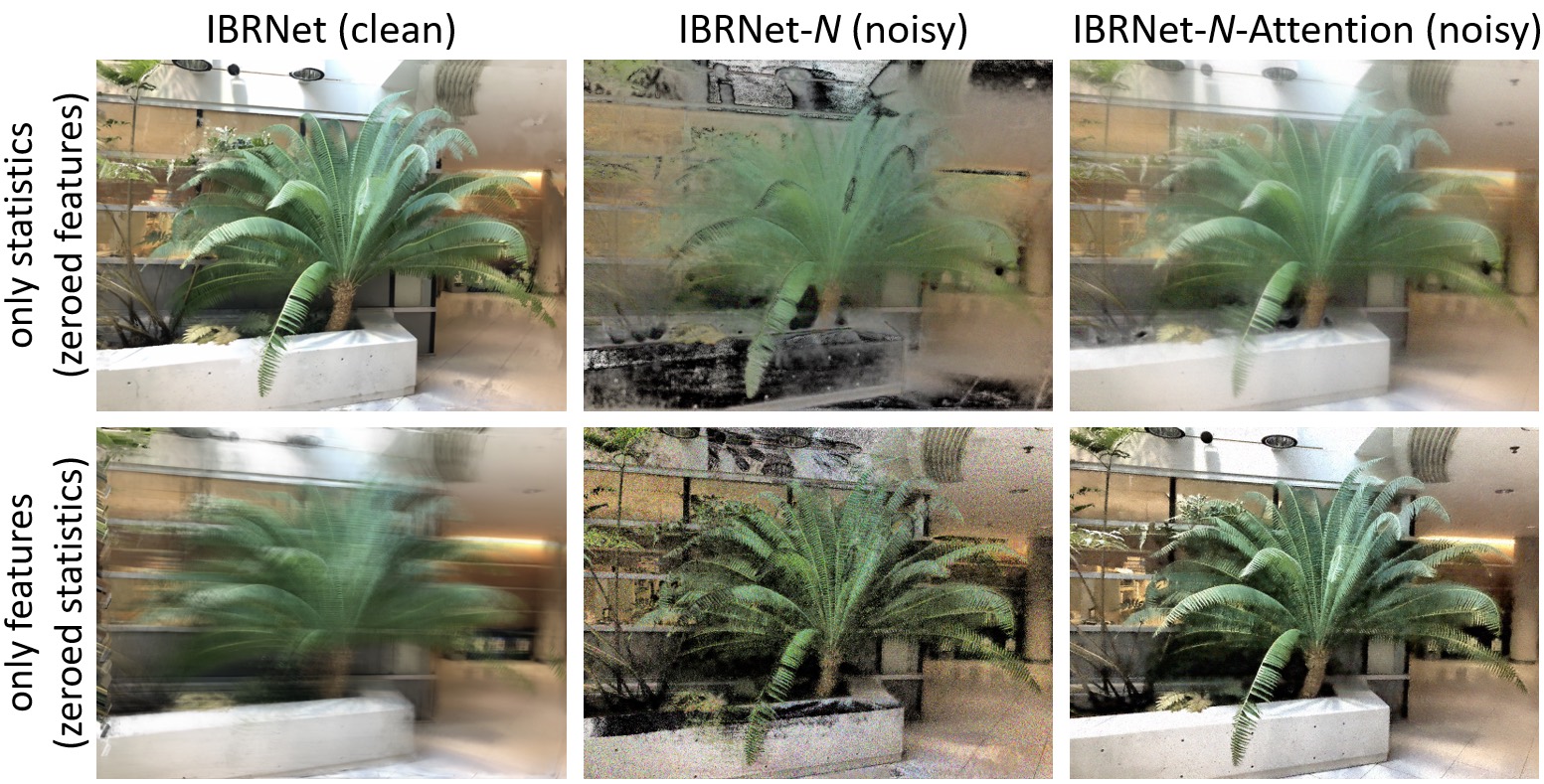}\vspace{-0.2cm}
    \caption{\textbf{Feature aggregation by the NAN attention module (Sec.~\ref{sec:feat_aggr})}: A demonstration of the importance and quality of individual features vs. joint statistics (mean, variance) in the clean vs. noisy regime, with and without our attention module. The \textbf{\textit{columns}} compare denoising performance of IBRNet on clean images [\textbf{left}] and IBRNet-$N$ on noisy images without \textbf{[middle]} and with [\textbf{right}] our attention module. The \textbf{\textit{rows}} compare the influence of the features statistics (mean, variance) [\textbf{top}] to that of the individual features [\textbf{bottom}]. 
    Please see text (in Sec \ref{sec:feat_aggr}) for a detailed discussion and interpretation of these results.} \vspace{-5pt}
    \label{fig:view_attn}
\end{figure}

The comparison in the case of the original IBRNet on clean images (Fig.~\ref{fig:view_attn} left column) clearly shows that the feature component is practically ignored by the network (the clean target image is omitted to avoid a trivial solution). 
This can be explained by the sterile conditions of lack of noise.
In this case, the variance of the features is indicative towards differentiating between (i) non-surface (i.e., free-space or occluded) 3D points, for which variance is typically high and (ii) visible surface points on which the rays intersect, for which variance is very low (no noise) and the mean well represents the color information (no noise).

As will be seen next, things change drastically with the presence of noise. In the middle column of Fig.~\ref{fig:view_attn}, we perform the same comparison on the inference of IBRNet-$N$ on noisy inputs (gain $16$). Here the variance and the mean cannot capture the density and color of the 3D point (as non-agreement due to noise or due to non-surface ray intersection can be confused), hence the original features are seen to be of greater significance to the downstream predictions. 

With this observation regarding the importance of the individual features in the noisy regime as well as the poor contribution of the mean and variance as summarizing statistics, we added a standard single 5-head self-attention transformer~\cite{vaswani2017attention} in order to capture the more intricate relations and statistics of the cross-view features (see \emph{II} in~Fig.~\ref{fig:density_color_scheme}).  Indeed, the right column of Fig.~\ref{fig:view_attn}, shows (especially considering the features-only bottom row) that the trained transformer provides a significant representation improvement to the vectors, resulting in a synthesized (denoised)  image with lower noise and better preservation of detail. We validate the contribution of the transformer in our  ablation study.

In addition, we add the estimated noise variances per-pixel per-color channel as additional features, concatenated to the global features (see Fig.~\ref{fig:density_color_scheme}). This helps the network to generalize over different noise levels without over-smoothing the results in lower noise levels~\cite{bhat2021deep, xia2020basis}.


\subsection{Noise-Aware Color Blending} \label{sec:color_blend}

In IBRNet~\cite{wang2021ibrnet} (and NeRF architectures in general), the output predicted color $c_m$ 
is obtained as a linear combination of the $N$ corresponding multi-view RGB-values, where the $N$ blending weights are predicted by an MLP at the end of the network (step 9 in Fig.~\ref{fig:density_color_scheme}). While this strategy is standard practice in image-based rendering, it is clearly insufficient for noise filtering, since only cross-view (single pixel) information is gathered, without any spatial extent.

We suggest replacing the length-$N$ blending vector, by a per-pixel per-view $k\times k$ spatial kernel of blending weights that is applied on the $k\times k$ spatial neighborhood of each projected pixel. 
Furthermore, we learn separate weights per color-channel. Thus, instead of outputting $N$ blending scalars, we output $N$ vectors, each of dimension $k^2\times 3$ (Fig.~\ref{fig:density_color_scheme}~\emph{III}). The separation per color channel is important especially in the case of signal-dependent noise, that can vary across the color channels. 
Prior to applying the blending, we normalize the weights per channel to sum to 1 by a softmax operation. 
In practice, we find the choice of $k=3$ to give a good balance between quality gain and increase in network complexity, and we add a simple bilateral filtering post-process, explained in Sec.~\ref{sec:results}. 
This approach of using 3D kernels for blending is closely related to the per-pixel 3D kernel prediction suggested by~\cite{mildenhall2018burst}. We demonstrate that this idea works well also in the context of NeRFs.

\begin{figure}[t!]
    \centering\vspace{-8pt}
    \includegraphics[width=0.99\columnwidth]{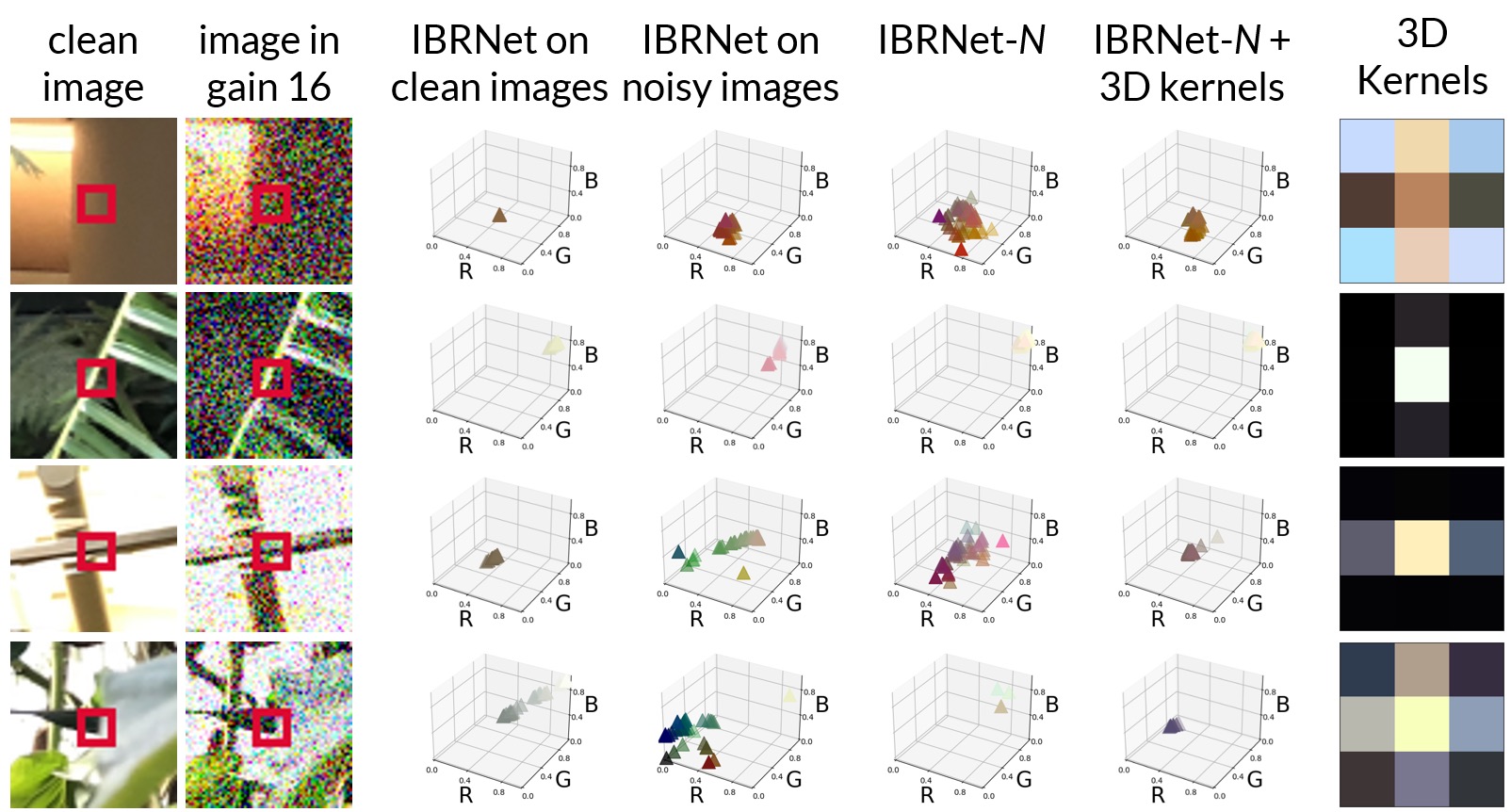}\vspace{-0.2cm}
    \caption{\textbf{Influence of the NAN 3D-blending kernels} (Sec.~\ref{sec:color_blend} and step \emph{III} in Fig.~\ref{fig:density_color_scheme}). 
    \textbf{[Left]}~Target pixels (clean and noisy with gain $16$). \textbf{[Center]}~Scatter plots in RGB-space of final predicted colors $c_m$ along the ray through the pixel (we show only colors that are more influential in the final integration along the ray - the ones at locations with above-average density $\rho_m$). These are shown for IBRNet on clean and noisy images, IBRNet-$N$ on noisy images, and IBRNet-$N$ with added 3D-blending kernels. \textbf{[Right]}~The learned 3D-kernels associated with the `true' pixel depth (we take the maximum-density sample), shown with RGB color-coding, scaled for clarity. See text for interpretation (Sec.~\ref{sec:color_blend}).  
    }\vspace{-5pt}
    \label{fig:kernels}
\end{figure}

An intuition regarding the importance of  the blending kernels is visualized in Fig.~\ref{fig:kernels}. We look at the final predicted colors along the ray  projected through several chosen example pixels (one per row). The RGB-space scatter-plots in the middle show the predicted dominant colors to be integrated over the ray. The added value of our 3D-blending kernels is clearly visible in the clean batch of colors, in comparison to the IBRNet-$N$ reference.  
In addition, our learned 3D-blending kernels (visualized on the right) can be seen to (i) have true spatial extent for an improved spatial support for denoising; (ii) have actual color values, which means that the channel separation was indeed exploited by the model.

\begin{figure*}
    \centering\vspace{-10pt}
    \includegraphics[width=1\textwidth]{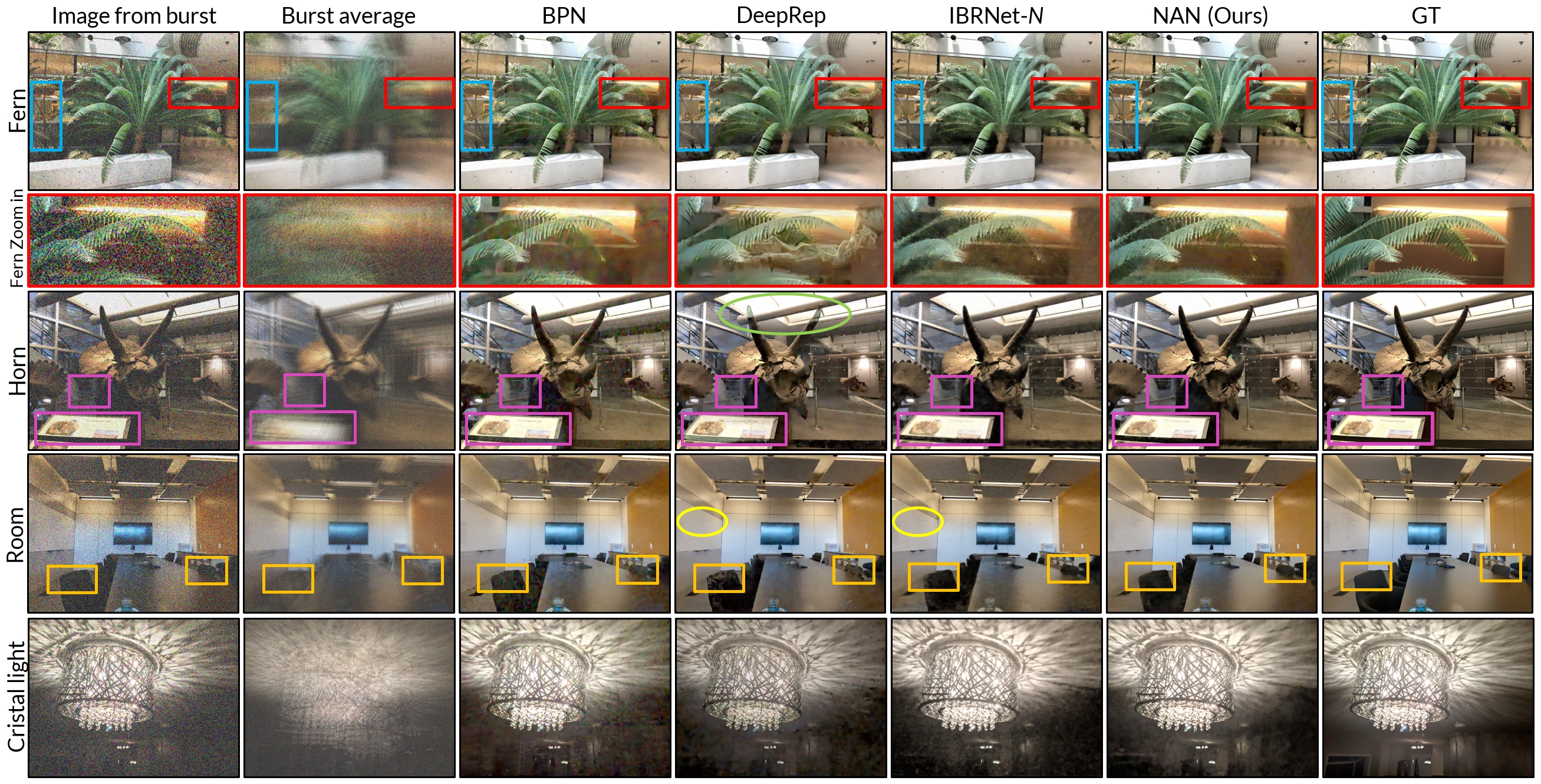} \vspace{-0.6cm}
    \caption{Comparison of results on the LLFF-$N$ dataset (rows 1-3), gain $20$, and a single scene (last row) with gain $16$ from the training set of~\cite{wang2021ibrnet} (not used in our training). We show a single noisy view, the burst average, results of BPN~\cite{xia2020basis}, DeepRep~\cite{bhat2021deep}, IBRNet-$N$, NAN, and the clean image. Interesting areas for comparison are marked by color rectangles. Color ellipses show artifacts (green) and missing detail (yellow) in competitor results. \textbf{The reader is encouraged to zoom-in.} }\vspace{-0.3cm}
    \label{fig:burst_denoising}
\end{figure*}

\section{Results} \label{sec:results}


\noindent\textbf{Implementation Details}. 
\hspace{-2pt}\textbf{(i)} \textbf{Generation of the LLFF-$N$} dataset is detailed in Sec.~\ref{sec:simple_baseline}.  
For training, we used $35$ scenes from LLFF~\cite{mildenhall2019local} and for testing - the test set from IBRNet~\cite{wang2021ibrnet} which consists of 5 scenes from~\cite{mildenhall2019local} and 3 scenes from ~\cite{mildenhall2020nerf}. 
\textbf{(ii) Loss:}
We use $l_1$ loss between the original (clean) pixel values and the predicted ones over a batch of random image projection rays $\mathcal{R}$. 
Choosing $l_1$ over the $l_2$ loss originally used in IBRNet was motivated by the findings in~\cite{zhao2016loss} regarding $l_1$ being generally more suitable for image restoration tasks, and indeed it improved results by an average of 0.4 dB in our case. 
%

\textbf{(iii) Training and evaluation:}
We train the NAN network on LLFF-$N$ for 255k iterations using the same training scheme as in~\cite{wang2021ibrnet}, taking around 2 days on a GeForce RTX 3090 GPU. We fix the resulting model and do not fine-tune it on the novel test scenes (which could possibly improve results at the cost of runtime, as suggested in~\cite{wang2021ibrnet}). 
The batch size is 512 rays, which includes processing of all views and points on each ray. Since the architecture of~\cite{xia2020basis} is limited to a fixed burst size of $8$ images, we decrease the burst sizes that were used in~\cite{wang2021ibrnet} and fix them to 8 for all methods. 
Note that the loss function is applied on linear space images, but the evaluation - on the reprocessed ones.
We train our network (and retrain~\cite{xia2020basis,  bhat2021deep}) on the extended noise-region (purple rectangle in Fig.~\ref{fig:noise_levels}) and test on several gain values in this range (black points in Fig.~\ref{fig:noise_levels}).

\textbf{(iv) Post-processing: }
We found that a simple bilateral filter~\cite{tomasi1998bilateral} nicely complements our method, as a fast post-processing stage at inference time only. It decreases noise levels in homogeneous regions, while not affecting textured ones. We also found this choice to give a good efficiency trade-off, instead of enlarging our blending kernels, which would cause an increase in runtime and memory footprint, both in training and inference.

\begin{figure} [!]
    \centering \vspace{-10pt}
    \includegraphics[width=0.94\columnwidth]{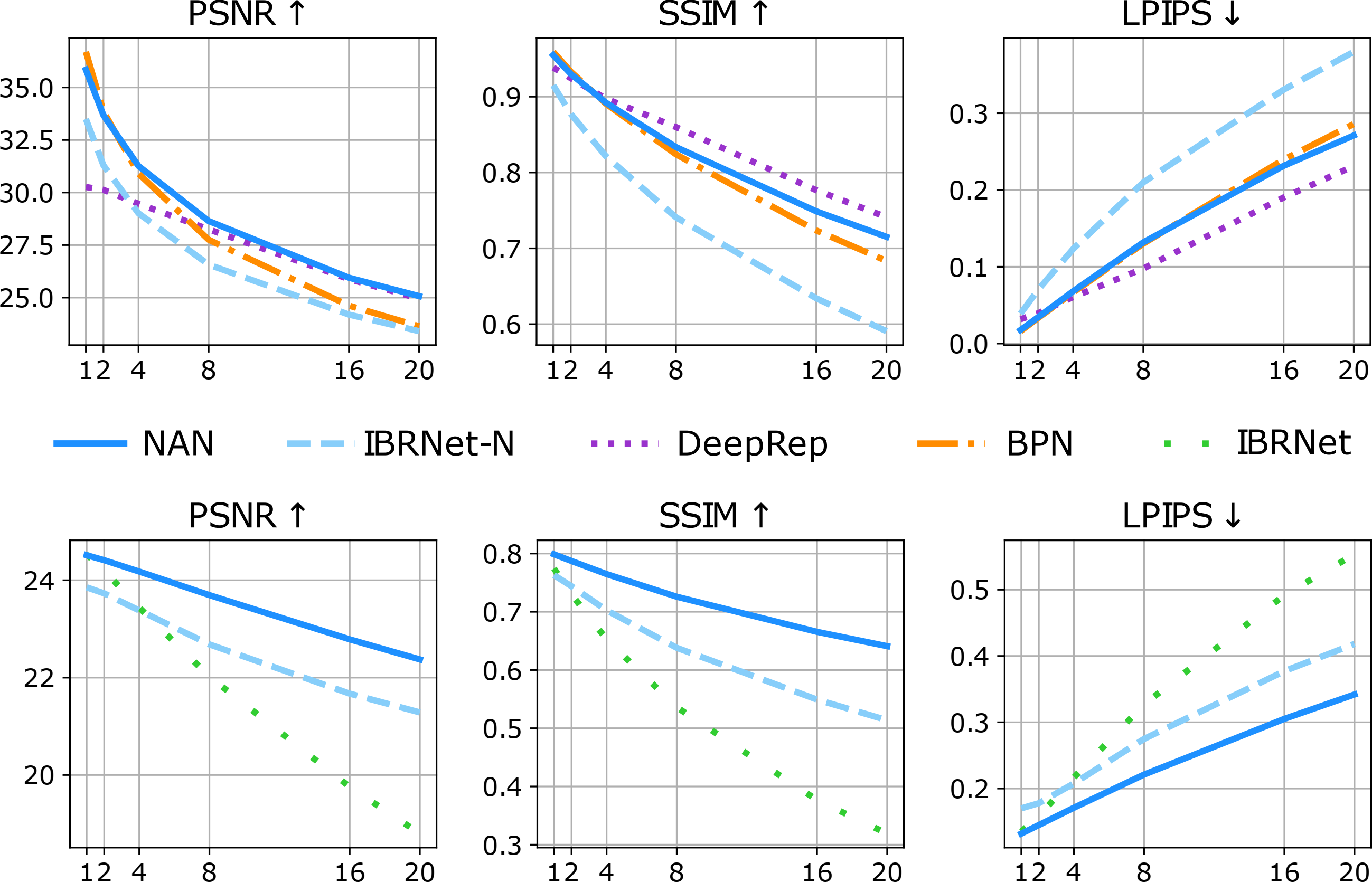}\vspace{-0.07in}
    \caption{\textbf{Comparison to SOTA burst denoising methods on LLFF-$N$ data}. Metrics for evaluation are PSNR, SSIM and LPIPS~\cite{zhang2018unreasonable} on gain levels of $1, 2, 4, 8, 16, 20$ ($x$-axis). \textbf{Top: Burst-Image Denoising} (using a noisy input target image); \textbf{Bottom: Novel-View Generation} (without input target view).}
    \label{fig:LLFF_graphs}\vspace{-2pt}
\end{figure}

\vspace{4pt}
\noindent\textbf{Results on LLFF-{\emph N}}.
As an evaluation set we use the $8$ real scenes that were used in~\cite{wang2021ibrnet}, with a total of $43$ bursts. 
We compare our results to the baseline IBRNet-\emph{N}, and to two recent SOTA burst denoising methods, BPN~\cite{xia2020basis} and DeepRep~\cite{bhat2021deep} on a range of gain levels.
Poses were extracted from the noisy images using COLMAP~\cite{schoenberger2016sfm}. 
Several examples results are shown in Fig.~\ref{fig:burst_denoising}, with color rectangles pointing to interesting areas. Fig.~\ref{fig:LLFF_graphs}~[top] summarizes the quantitative results in terms of PSNR, SSIM and LPIPS~\cite{zhang2018unreasonable}.
In PSNR, interestingly, there is a ``scissor-like" behavior between BPN and DeepRep, where BPN/DeepRep is better in lower/higher noise. In contrast, NAN consistently outperform the others.
In terms of SSIM and LPIPS we improve on BPN, especially in higher noise levels, and are competitive with DeepRep. 
While DeepRep's SSIM scores are better, we notice that its results contain many qualitative inconsistencies due to artifacts, over-smoothing and loss of detail, as can be seen in crop areas in Fig.~\ref{fig:burst_denoising}.  

\vspace{4pt}
\noindent\textbf{Novel-View Results.} Additionally, we  test novel-view generation from noisy images by retraining the network to conduct inference without receiving the target image. Quantitative results in Fig.~\ref{fig:LLFF_graphs}~[bottom] show consistent improvements across noise-levels and qualitative results are presented in the NAN website.

\vspace{4pt}
\noindent\textbf{Real-World Results.}  Fig.~\ref{fig:real_world}  depicts real-world results in two challenging low-light scenes, both photographed with a Google Pixel~4 phone.
Both image sets contain $8$ frames that were saved in RAW format. On both image sets, NAN outperforms BPN~\cite{xia2020basis} and DeepRep~\cite{bhat2021deep}, with less artifacts and sharper results with a finer level of detail. Additional results are presented in the NAN web-page. %
The input to the algorithms are the linear images scaled to the range of $[0,1]$ to match the network training, with noise parameters extracted from the EXIF file and scaled using the brightness scale of the images. Results are post-processed for display 
 (see interesting details marked on the images). The  BPN~\cite{xia2020basis} results have visible square artifacts stemming from the original implementation that is limited in resolution. 

\begin{figure}[t]
\vspace{2pt}
\setlength{\tabcolsep}{5pt} 
\begin{tabular}{cc}
\hspace{-7pt}
\begin{minipage}{.425\columnwidth} \vspace{-12pt}
    \caption{Denoising error as a function of pixel displacement in gains $1$, $8$ and $20$. We calculated the average displacement for each object point using optical-flow~\cite{revaud2015epicflow}. 
    [Bottom]~Histogram of pixel displacements in our test set. Pixel count is in log space. [Top]~We calculated the average MSE reconstruction error per displacement. 
    See text for interpretation.}
\label{fig:psnr_of}\vspace{-15pt}
\end{minipage}
&
\begin{minipage}{.9\columnwidth}
\vspace{-10pt}
\includegraphics[width=.585\columnwidth]{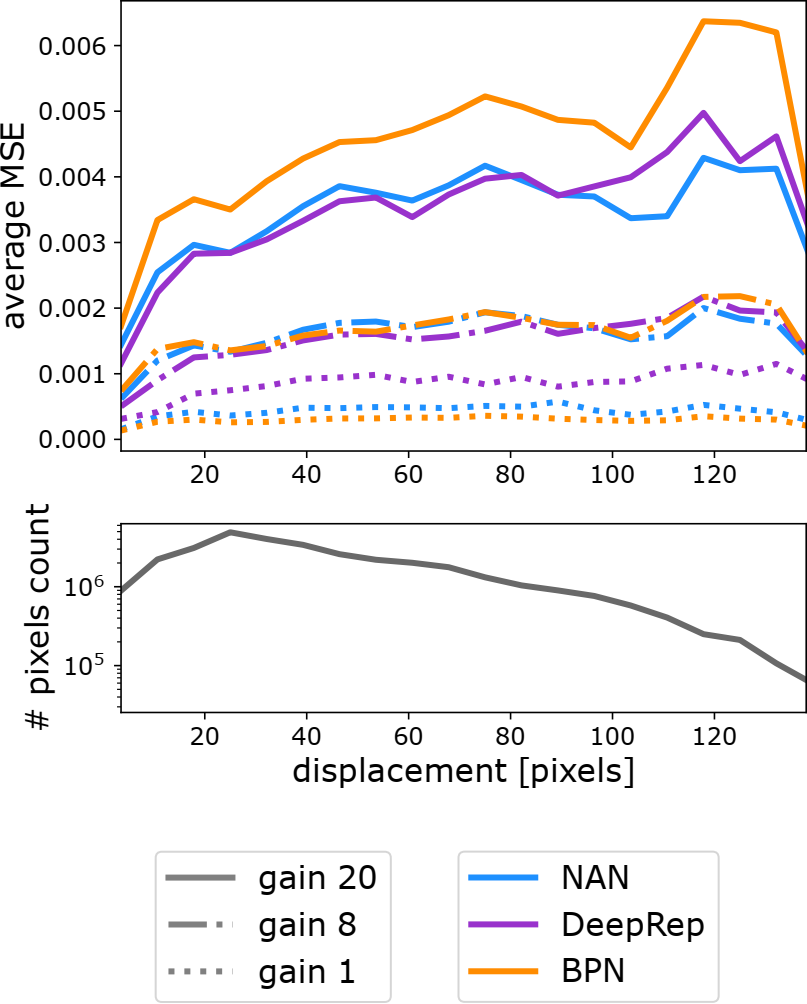}
\end{minipage}
\\
\end{tabular}
         \\ \vspace{-16pt} \\
\end{figure}

\vspace{4pt}
\noindent\textbf{Robustness to Motion.} For each point of each target image in LLFF-$N$ we calculated the average 2D displacement from each of the other frames, using EpicFlow~\cite{revaud2015epicflow}. In Fig.~\ref{fig:psnr_of} we compared average MSE results, for gain levels $1$, $8$ and $20$, as a function of the calculated average displacement. The distribution of disparities (bottom) shows presence of large motion that needs to be dealt with.
It is noticeable that NAN outperforms the competitors in the challenging regime of high noise and large motion and its performance is the least affected by motion magnitude. 

\vspace{4pt}
\noindent\textbf{Ablation Study. }\label{sec:ablation} 
We ablate the performance of NAN on the LLFF-$N$ data with gain $16$, with results in Table~\ref{tab:ablation} demonstrating the efficacy of our proposed additions. 
We also calculated the error of the predicted depth map with respect to the depth map generated by~\cite{wang2021ibrnet} from the clean images. 
We see a steady improvement from each of the components, with the 3D-blending being most significant - demonstrating that incorporating spatial awareness within the network improves 3D understanding, which might be a key factor in burst-denoising. 
We provide further ablations, regarding loss functions and kernel sizes in the NAN web-page.
%

\begin{table}[t]
\centering \vspace{2pt}
\includegraphics[width=0.99\columnwidth]{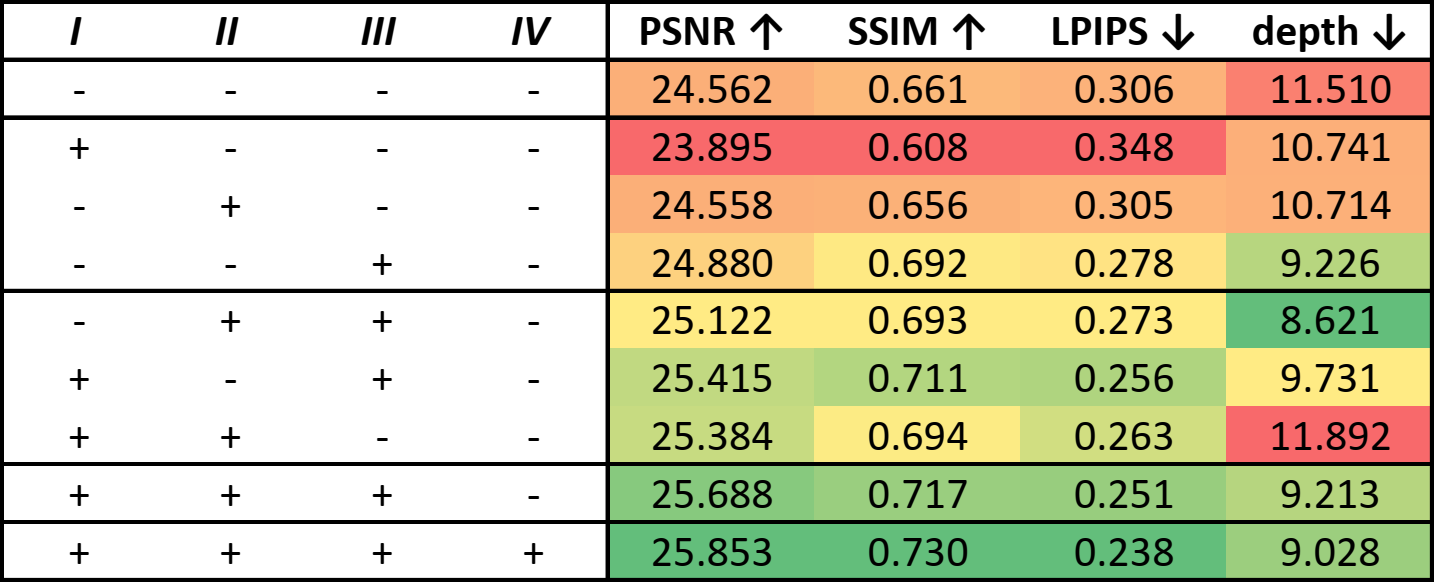}
\vspace{-0.07in}
\caption{\textbf{Ablation study}. Trained on the high noise range rectangle, evaluated on gain $16$. \emph{I}- pre-net, \emph{II} - feature aggregation, \emph{III} - 3D-blending, \emph{IV} - noise parameters. Depth error is average MSE.}
\label{tab:ablation} \vspace{-8pt}
\end{table}

\section{Discussion}\vspace{-2pt}

We demonstrated that the NeRF solution can serve as a powerful burst denoising paradigm for challenging sets containing large motion, parallax effects and high noise. Implicit consideration of the underlying scene geometry regularizes the displacement and imposes an excellent prior. It enables exploiting spatial and inter-frame information to infer the clean scene appearance. We show denoising of challenging motion in the order of $100$ pixels, including occlusions, under severe noise, higher than previously tested.

Limitations of NAN lead to our future plans: Currently we calculate the poses in pre-processing, while recent methods have incorporated pose estimation into the NeRF itself. We plan doing so, and hypothesize that it will enable coping with even higher noise levels, that currently prohibit separate pose estimation. The runtime of NAN is its main drawback - a common issue with NeRFs. Nevertheless, recent advancements have shown impressive results in speeding up and parallelizing radiance field calculations, which we are sure our framework can benefit from. We believe that with these future developments, burst denoising using NeRFs can result in a very compelling framework for challenging burst-denoising inputs.

\vspace{4pt}
\noindent {\small \textbf{Acknowledgements.} The research was funded by Israel Science Foundation grant $\#680/18$ and the Israeli Ministry of Science and Technology grant $\#3-15621$, the  Leona M. and Harry B. Helmsley Charitable Trust, the Maurice Hatter Foundation. 
We thank the Interuniversity Institute for Marine Sciences of Eilat for making their facilities available to us; and Deborah Steinberger-Levy, Opher Bar Nathan, and Amit Peleg for help with experiments.} 




{\small
\bibliographystyle{ieee_fullname}
\bibliography{egbib}
}

\end{document}